\documentclass[10pt,twocolumn,letterpaper]{article}

\usepackage{cvpr}              

\usepackage{graphicx}
\usepackage{amsmath}
\usepackage{amssymb}
\usepackage{booktabs}
\usepackage{multirow}
\usepackage[accsupp]{axessibility}  

\usepackage[pagebackref,breaklinks,colorlinks]{hyperref}

\def\mathbi#1{\textbf{\em #1}}

\usepackage[capitalize]{cleveref}
\crefname{section}{Sec.}{Secs.}
\Crefname{section}{Section}{Sections}
\Crefname{table}{Table}{Tables}
\crefname{table}{Tab.}{Tabs.}

\def\cvprPaperID{7185} 
\def\confName{CVPR}
\def\confYear{2023}

\begin{document}


\title{NAR-Former: Neural Architecture Representation Learning towards Holistic Attributes Prediction}
\author{Yun Yi$^1$ \quad Haokui Zhang$^{2,3,*}$ \quad Wenze Hu$^2$ \quad Nannan Wang$^{1, *}$ \quad Xiaoyu Wang$^2$\\
$^1$Xidian University \quad $^2$Intellifusion \quad $^3$Harbin Institute of Technology, Shenzhen\\
{\tt\small yuny220@163.com \quad hkzhang1991@mail.nwpu.edu.cn}\\
{\tt\small windsor.hwu@gmail.com \quad nnwang@xidian.edu.cn \quad fanghuaxue@gmail.com}
}
\maketitle
\footnotetext{$^1$This work was done while Yun Yi was an intern at Intellifusion.}
\footnotetext{$^*$Corresponding authors.}

\begin{abstract}
With the wide and deep adoption of deep learning models in real applications, there is an increasing need to model and learn the representations of the neural networks themselves. These models can be used to estimate attributes of different neural network architectures such as the accuracy and latency, without running the actual training or inference tasks. In this paper, we propose a neural architecture representation model that can be used to estimate these attributes holistically. Specifically, we first propose a simple and effective tokenizer to encode both the operation and topology information of a neural network into a single sequence. Then, we design a multi-stage fusion transformer to build a compact vector representation from the converted sequence. For efficient model training, we further propose an information flow consistency augmentation and correspondingly design an architecture consistency loss, which brings more benefits with less augmentation samples compared with previous random augmentation strategies. Experiment results on NAS-Bench-101, NAS-Bench-201, DARTS search space and NNLQP show that our proposed framework can be used to predict the aforementioned latency and accuracy attributes of both cell architectures and whole deep neural networks, and achieves promising performance. Code is available at \href{https://github.com/yuny220/NAR-Former}{https://github.com/yuny220/NAR-Former}.


\end{abstract}

\section{Introduction}
\label{introduction}

As an ever increasing variety of deep neural network models are widely adopted in academic research and real applications, neural architecture representation is emerging as an universal need to predict model attributes holistically. For example, modern neural architecture search (NAS) methods can depend on the neural architecture representation to build good model accuracy predictors\cite{luo2018neural,wang2019alphax,shi2020bridging,ning2020generic,lu2021tnasp,chen2021not}, which estimate model accuracies without running the expensive training procedure. In order to find faster execution graphs when deploying models to neural network accelerators,  neural network compilers \cite{chen2018tvm, liu2022nnlqp} need it to build network latency predictors, which estimate the real time cost without running it on the corresponding real hardware. As a straightforward approach to solve these holistic prediction tasks, a neural architecture representation model should be built to take the symbolic description of the network as input, and generate its numerical representation which can be easily used with existing modeling tools to perform the desired downstream tasks.

There are some neural architecture representation models proposed in solving the individual tasks mentioned above, but we find they are rather application specific and have obvious drawbacks when used in new tasks.  For example, the early MLP, LSTM based accuracy prediction approaches\cite{deng2017peephole,liu2018progressive,luo2018neural,wang2019alphax} improve the efficiency and performance of NAS, while there is a limitation of prediction performance resulting from the the nonexistent or implicit topology encoding.
Some later proposed GCN based methods\cite{chen2021contrastive,li2020neural,wen2020neural} achieve better performance of accuracy prediction than the above methods.  However, due to the use
of adjacency matrix, the encoding dimension of these methods
scales quadratically with the depth of the input architecture, 
making them difficult to model large networks.
NNLQP\cite{liu2022nnlqp} implements latency prediction at the model level, but GNN-based encoding scheme makes it inadequate in modeling long range interactions between nodes.

Inspired by the progress in natural language understanding, our network uses a hand designed yet general and extensible token encoding approach to encode the topology information and key parameters of the input neural network. 
Specifically, we design a generic real value tokenizer to encode neural network nodes' operation type, location, and their inputs into vectors using the positional embedding approach that is widely used in transformers\cite{vaswani2017attention} and NERF\cite{mildenhall2021nerf}.
This fixed length node encoding scheme makes the network encoding scale linearly with the size of the input network rather than quadratically as in models that rely on taking adjacency matrix as input.


Based on the tokenized input sequence, we design a transformer based model to further encode and fuse the network descriptions to generate a compact representation, \eg a fixed length feature vector. Therefore, long range dependencies between nodes can be established by transformer. Besides the conventional multi-head self attention based transformers, we propose to use a multi-stage fusion transformer structure at the deep stages of the model to further fuse and refine the representation in a cascade manner.

For the network representation learning task, we also propose a data augmentation method called information flow consistency augmentation. The augmentation permutes the encoding order of nodes in the original network under our specified conditions without changing the structure of the network. We find it empirically improves the performance of our method for downstream tasks.

The main contributions of this paper can be summarized as the following points:
\begin{enumerate}    
    \item We propose a simple and effective neural network encoding approach which tokenizes both operation and topology information of a neural network node into a sequence. When it is used to encode the entire network, the tokenized network encoding scheme scales better than adjacency matrix based ones, and builds the foundation for seamlessly using transformer structures for network representation learning. 

    \item We design a multi-stage fusion transformer to learn feature representations. Benefiting from the proposed tokenizer, a concise pure transformer based neural architecture representation learning framework (NAR-Former) is proposed for the first time. Our NAR-Former makes full use of the capacity of transformers in handling sequence inputs, and gets promising performance.  

    \item To facilitate efficient model training, we propose an information flow consistency augmentation and correspondingly design an architecture consistency loss, which brings more benefits with less augmentation samples compared with the existing random augmentation strategy.

    
    
  
\end{enumerate}

We conduct extensive experiments on accuracy prediction, neural architecture search as well as latency prediction. Experiments demonstrate the effectiveness of the proposed representation model on processing both cell leveled network components and whole deep neural networks. Specifically, our accuracy predictor based on this representation achieves highly competitive accuracy performance on cell-based structures in NAS-Bench-101\cite{ying2019bench} and NAS-Bench-201\cite{dong2020bench} datasets. Compared with other predictor-based NAS methods\cite{lu2021tnasp,ning2020generic}, we efficiently find the architecture with 97.52\% accuracy in DARTS\cite{liu2018darts} by only querying 100 neural architectures. We also conduct latency prediction experiments on neural networks deeper than 200 layers to demonstrate the universality of our model.


\begin{figure*}[ht]
    \centering
    \includegraphics[width=\linewidth]{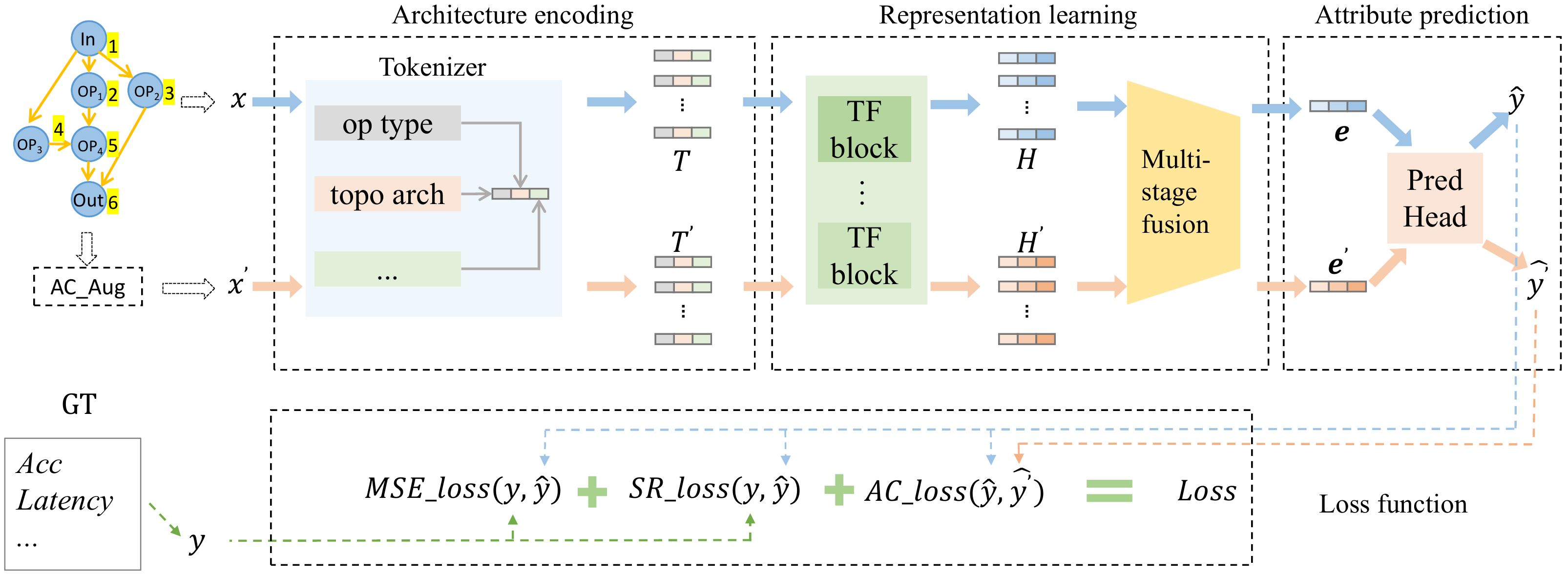}
    \caption{Overview of our NAR-Former. 
    We first encode an input architecture $x$ to a pure sequence $T$ with a proposed tokenizer. A multi-stage fusion transformer is designed to learn a vector representation  $\mathbi{e}$ from $T$. $x'$(optional) is an augmented architecture of $x$ generated by our information flow consistency augmentation. The bottom part shows the loss function used in this paper. SR\_loss is designed for learning more accurate sequence ranking. AC\_loss is a proposed architecture consistency loss.}
    \label{fig:overall}
\end{figure*}

\section{Related Work}
\label{related_work}

\subsection{Representation of Neural Architectures}

Neural architecture representation learning is essential for implementing downstream tasks (\eg accuracy prediction and latency prediction). 
Many accuracy predictors model the input architectures by adopting LSTM\cite{luo2018neural,liu2018progressive,deng2017peephole} and MLP\cite{liu2018progressive,white2021bananas}, which are used to accelerate NAS. They usually suffer from a limited accuracy because of the insufficient encoding of the topology structure. For example, 
NAO\cite{luo2018neural} directly encodes the whole architecture with a flatted vector followed by an embedding layer.
Recently, many predictors explore how to use GCNs\cite{chen2021contrastive,shi2020bridging,li2020neural,wen2020neural} to model the graph format data(\eg adjacency matrix and node features) of architectures. Neural Predictor\cite{wen2020neural} encodes each node in architecture with the mean of two modified GCN layers, which use the adjacency matrix and the transpose of adjacency matrix to propagate information, respectively. When the input architecture becomes deeper, there is a quadratically increase of encoding dimension in these methods caused by the adjacency matrix.
There are methods \cite{xu2021renas, chen2021not} that use other models for representation learning. HOP\cite{chen2021not} proposes an adaptive attention module to model the relative significance of different network nodes for the first time, and achieved attractive prediction performance. However, the lacking consideration of the networks' overall characteristics leads to the fact that these methods are not suitable for extending to whole networks. 

Latency prediction often needs to take the information of whole network into consideration when modeling the input architecture. TPU-Performance\cite{kaufman2019learned,kaufman2021learned} and nn-Meter\cite{zhang2021nn} encode architectures by dividing a whole network inference pass into multiple kernels, and calculate the sum of kernels' latency. In real applications, the hardware graph fusion may bring a gap between the sum of kernel latencies and the real latency of the whole network. To alleviate this problem, NNLQP \cite{liu2022nnlqp} proposes a unified GNN-based representation model for various neural architectures and predicts the latency in model level. 
But the model in NNLQP\cite{liu2022nnlqp} is rather preliminary since it generates architecture representations by directly summing up the features of all nodes and is not good at capturing mode sequence information.

\subsection{Transformers}
The transformer model \cite{vaswani2017attention}, originally proposed in natural language processing(NLP), proposes to draw global dependencies and realize parallel computation using  multi-head self-attention mechanism. Because of the outstanding performance in NLP \cite{radford2018improving,devlin2018bert} and its domain agnostic network structure, transformers are quickly being used in many other fields\cite{liu2021swin,dosovitskiy2020image,carion2020end}. TNASP\cite{lu2021tnasp} adopts transformer to encode neural network for the first time, which uses the MLP transform of the Laplacian matrix as the position encoding. It achieves better performance compared to some existing methods, which confirms the advantage of transformer for architecture representation learning. 
However, the Laplacian matrix of the network with different depths needs to be filled to a preset shape when calculating  the position encoding, which may introduce unwanted information. We argue that the ability to handle sequences of variable length and to model long range dependency makes transformer an ideal backbone for neural architecture representation learning, if the given structure is mapped into a single sequence as its input. 
\section{The Proposed Method}
\label{method}

\subsection{Overall Framework}

\cref{fig:overall} shows the overall framework of our proposed NAR-Former. Firstly, the input cell or network $x$ that has $N$ nodes will be encoded to a token sequence $T\in\mathbb{R}^{(N+2) \times D}$ with a proposed tokenizer(\cref{sec:NAE}). Then a  multi-stage fusion transformer(\cref{sec:MS-fusion}) is designed to transform $T$ to a one-token feature representation $\mathbi{e}\in\mathbb{R}^{1 \times D}$. At the last stage, attribute value $\widehat{y}$ will be predicted using a prediction head that consists of fully connected layers and activation layers. After training with $M$ different architecture-attribute pairs $\left \{ (x_{i}, y_{i}) \right \}_{i=1}^M$, NAR-Former can predict the attribute of an unseen architecture by only one forward inference pass, regardless of the number of operations.

We propose an information flow consistency augmentation method to further improve the accuracy of attribute prediction(\cref{sec:AC-Aug}), which is labeled ``AC\_Aug'' in \cref{fig:overall}. $x'$ is an architecture obtained by augmenting $x$, and $\widehat{y'}$ is its prediction. Corresponding to the proposed augmentation, an architecture consistency loss is designed(\cref{sec:loss}).

The bottom part of \cref{fig:overall} illustrates the loss function(\cref{sec:loss}) we used, consisting of MSE loss and two designed loss, which are a loss related to sequence ranking(SR\_loss) and architecture consistency loss(AC\_loss).

\subsection{Architecture Encoding}\label{sec:NAE}
\begin{figure}[ht]
    \centering
    \includegraphics[width=\linewidth]{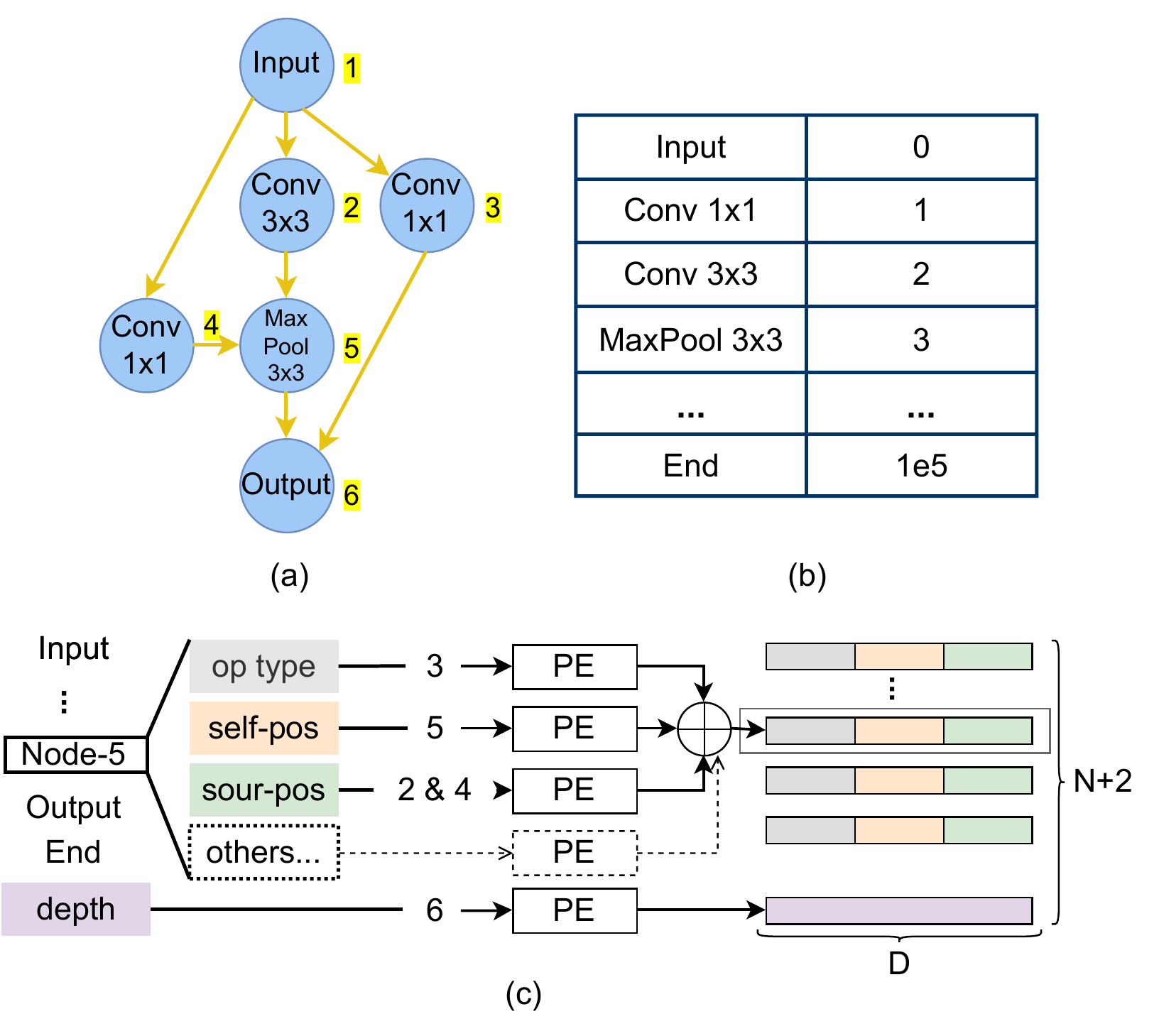}
    \caption{(a) An example of architecture with 6 operations($N=6$). (b) Conversion table from operation categories to indexes. (c) Encoding scheme of our tokenizer.}
    \label{fig:embedding}
\end{figure}


In order to avoid information loss of the input architecture and use transformer seamlessly, we propose a tokenizer to encode operations and their connections of a neural network into a sequence of tokens. 
When given a neural network, any operation that exists in this network can be considered as a node.
Our tokenizer can encode the operator and topological structure of the input architecture $x$ that has $N$ operations into a $N+2$ token sequence $T$, where each token $\mathbi{t}$ is a $D$ dimension vector.
\begin{equation}\label{eqn::concat}
    T = \left(\mathbi{t}_1, \mathbi{t}_2, \cdots, \mathbi{t}_{N}, \mathbi{t}_{\mathrm{End}}, \mathbi{t}_{\mathrm{Dep}}\right).
\end{equation}
The first $N$ tokens are the tokens for the $N$ nodes and their connections. $\mathbi{t}_{\mathrm{End}}$ is an end token that signifies the end of network encoding, and the last token $\mathbi{t}_{\mathrm{Dep}}$ encodes the depth of the input network. Each token $\mathbi{t}$ in $\mathbi{t}_i(i\in(1,2,\cdots,N,\mathrm{End}))$  is further decomposed into 3 sub vectors: 
\begin{equation}
    \mathbi{t}_i = \left( \mathbi{t}_i^{\mathrm{op}}, \mathbi{t}_i^{\mathrm{self}}, \mathbi{t}_i^{\mathrm{sour}} \right),
\end{equation}
which are the encodings of operator type $\mathbi{t}_i^{\mathrm{op}}$, its location $\mathbi{t}_i^{\mathrm{self}}$ and the summarized location of its source nodes $\mathbi{t}_i^{\mathrm{sour}}$. 



\noindent \textbf{Generic real number tokenizer.} The information to be encoded above can all be represented first in the form of real numbers. Inspired by the encoders used in NERF \cite{mildenhall2021nerf}, we propose the following encoding function $f$ that maps a real number $p$ to a vector in $\mathbb{R}^{2L}$:
\begin{equation}
\begin{split}
\label{eq:pe}
    f(p) = \left[\sin(b_1p\pi), \cos(b_1p\pi), \cdots, 
            \right.
            \\
            \left.
            \sin(b_Lp\pi), \cos(b_Lp\pi) \right].
\end{split}
\end{equation}.

Different from the geometric frequencies $(2^0, 2^1, \cdots, 2^{L-2}, 2^{L-1})$ used in \cite{mildenhall2021nerf}, we obtain the frequencies by linear interpolation in  range $[2^0, 2^{L-1}]$:
\begin{equation}
    \mathbi{b} = (2^0, 2^0+s, \cdots, 2^0+(L-2)s, 2^{L-1}),
\end{equation}
where,
\begin{equation}
    s=\frac{2^{L-1}-2^0}{L-1}.
\end{equation}
Depending on the range of the encoded real number, the length parameter $L$ can be specified on demand.

\noindent \textbf{Encoding for Operation Type Information.}
As is illustrated in \cref{fig:embedding}(b), we can use a conversion table to map the operator types into their indexes. The integer index can then be converted to a vector using the Generic real number tokenizer. Denoting $\mathcal{C}(v_i)$ as the operation type index of node $v_i$, the operation type encoding can be defined as
\begin{equation}
    \mathbi{t}_i^{\mathrm{op}} = f_{\mathrm{op}}(\mathcal{C}(v_i)).
\end{equation}

 The operations that can be encoded here includes, but are not limited to, basic mathematical operation(\eg addition, multiplication), convolution operation, normalization operation, even channel operation(\eg concatenation). 


\noindent \textbf{Encoding for Topology Information.} 
The encoding of topology structure can be decomposed to encoding of the information flow in each node of the network. For each node, we encode the location of the node itself and its input nodes in the network structure $x$.
Location of the node $v_i$ itself (``self-pos'' in \cref{fig:embedding}(c)) is encoded as
\begin{equation}
    \mathbi{t}_i^{\mathrm{self}} = f_{\mathrm{self}}(i).
\end{equation}

The locations of its source nodes are encoded into one vector as the sum of all source location encodings (``sour-pos'' in \cref{fig:embedding}(c)):
\begin{equation}
    \mathbi{t}_i^{\mathrm{sour}} = \sum_{j\in \mathcal{P}(v_i)}f_{\mathrm{sour}}(j),
\end{equation}
where $\mathcal{P}(v_i)$ is the set of locations of all the node's precursors. Then the concatenation of $\mathbi{t}_i^{\mathrm{self}}$ and $\mathbi{t}_i^{\mathrm{sour}}$ along the feature dimension is used to represent topology information. 



\noindent \textbf{Special Tokens} For $v_0$, which has no precursor node, we set $\mathcal{P}(v_0)$ equals to $\left\{ -1 \right\}$. For the end token, both the self position encoding and source position encoding can be encodings of a value that is unlikely to appear in the architecture code, such as $1\times 10^5$ in our experiments.

Considering that the depth of a network is also an important factor affecting its attributes, an additional depth token is encoded by using \cref{eq:pe} to map the depth $N$ into a $D$-length vector:
\begin{equation}
    \mathbi{t}_{\mathrm{Dep}} = \mathrm{ReLU}(\mathrm{FC}(f_{\mathrm{Dep}}(N))),
\end{equation}
where FC denotes a fully connected layer. 

\noindent \textbf{Extensibility.} 
Following the approach to encode operator types and locations, the token $\mathbi{t}$ in \cref{eqn::concat} can be naturally extended to further encode other information necessary for downstream tasks. For example, we further encode key parameters and output shape of each node by concatenating their encodings from real value tokenizers in our network latency prediction experiments (\cref{sec:NNLQP}). Due to space limitations, we will introduce the content of these additional encoded information in the supplementary material.

\subsection{Multi-Stage Fusion Transformer}\label{sec:MS-fusion}
\begin{figure}[ht]
    \centering
    \includegraphics[width=0.8\linewidth]{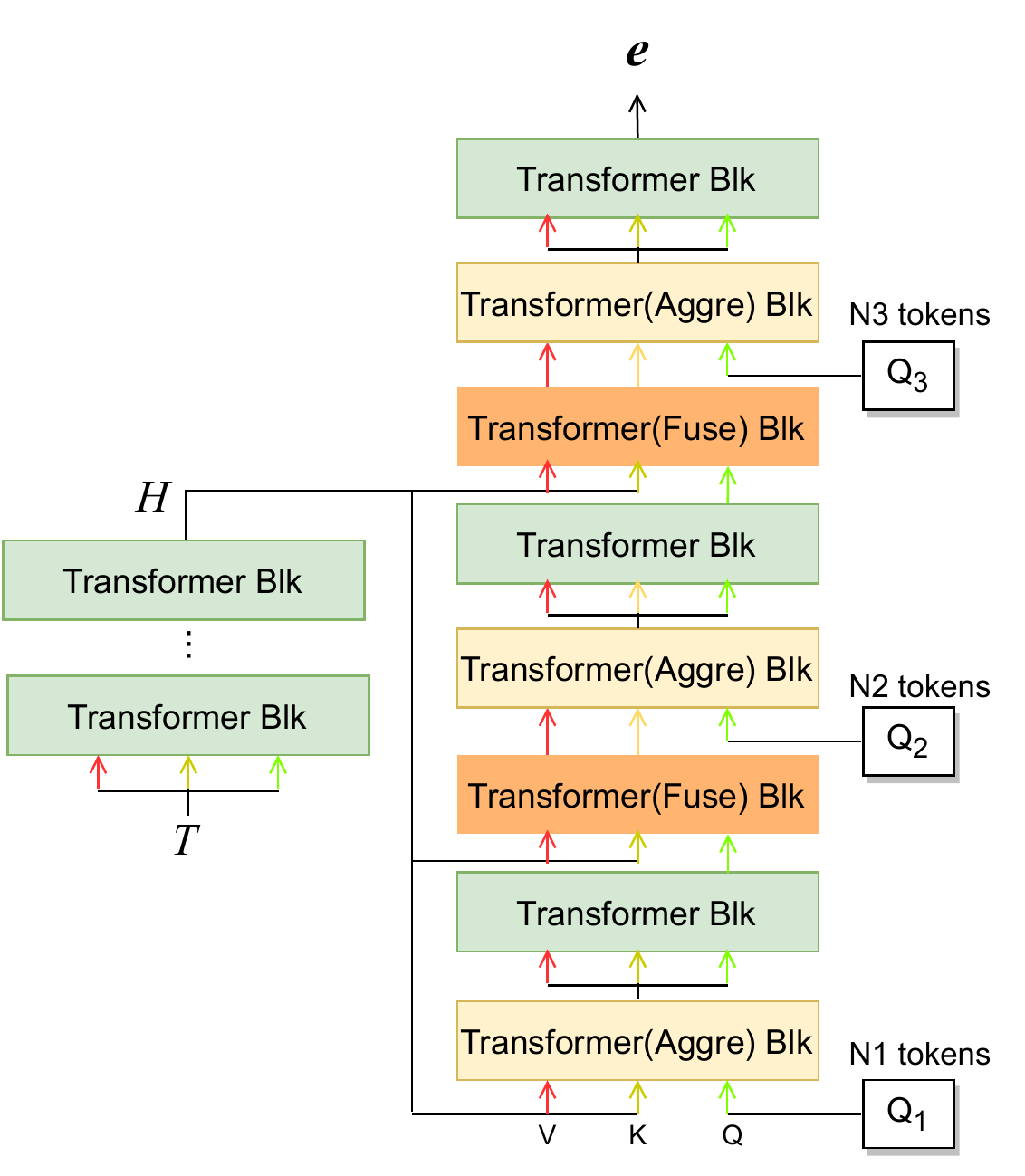}
    \caption{Multi-Stage Fusion Transformer. Token sequence $T$ is first transformed into feature map $H$ by standard transformer blocks, and then gradually fused into a one-token feature vector $\mathbi{e}$.}
    \label{fig:multi-stage}
\end{figure}

Our transformer-based feature representation learning model consists of normal transformer blocks and the multi-stage fusion part as shown in \cref{fig:multi-stage}. The red, yellow, and green arrows from left to right indicate the value, key, and query inputs for each transformer block, respectively. At the first stage, normal transformer blocks are used to further transform the tokens in $T$ and learn an intermediate token sequence $H$ through global interactions. 

The multi-stage fusion part of our model is consisted of three types transformer blocks which we call aggregation blocks, standard transformer blocks and fusion blocks respectively. The standard blocks are optional depending on the requirements of model size and performance. In the aggregation block, we use a learnable parameter matrix $Q_k$ as the query, and use $H$(only in the first aggregation blok) or the previous fuse block's output as both key and value of the transformer input. Depending on the level parameter $k$, the number of query tokens(denoted as $N_k$) in $Q_k$ is different, which helps shrink (thus aggregate information) the number of output tokens as the level goes up. The fuse block uses the output of a dependent standard transformer as the query, and uses the $H$ as key and value of the transformer input. Specifically, the number of query tokens in the last aggregation block is shrinked to 1. Thus, the output is a single feature vector regardless of the size of the input network $x$.

\subsection{Information Flow Consistency Augmentation}\label{sec:AC-Aug}
\begin{figure}[ht]
    \centering
    \includegraphics[width=\linewidth]{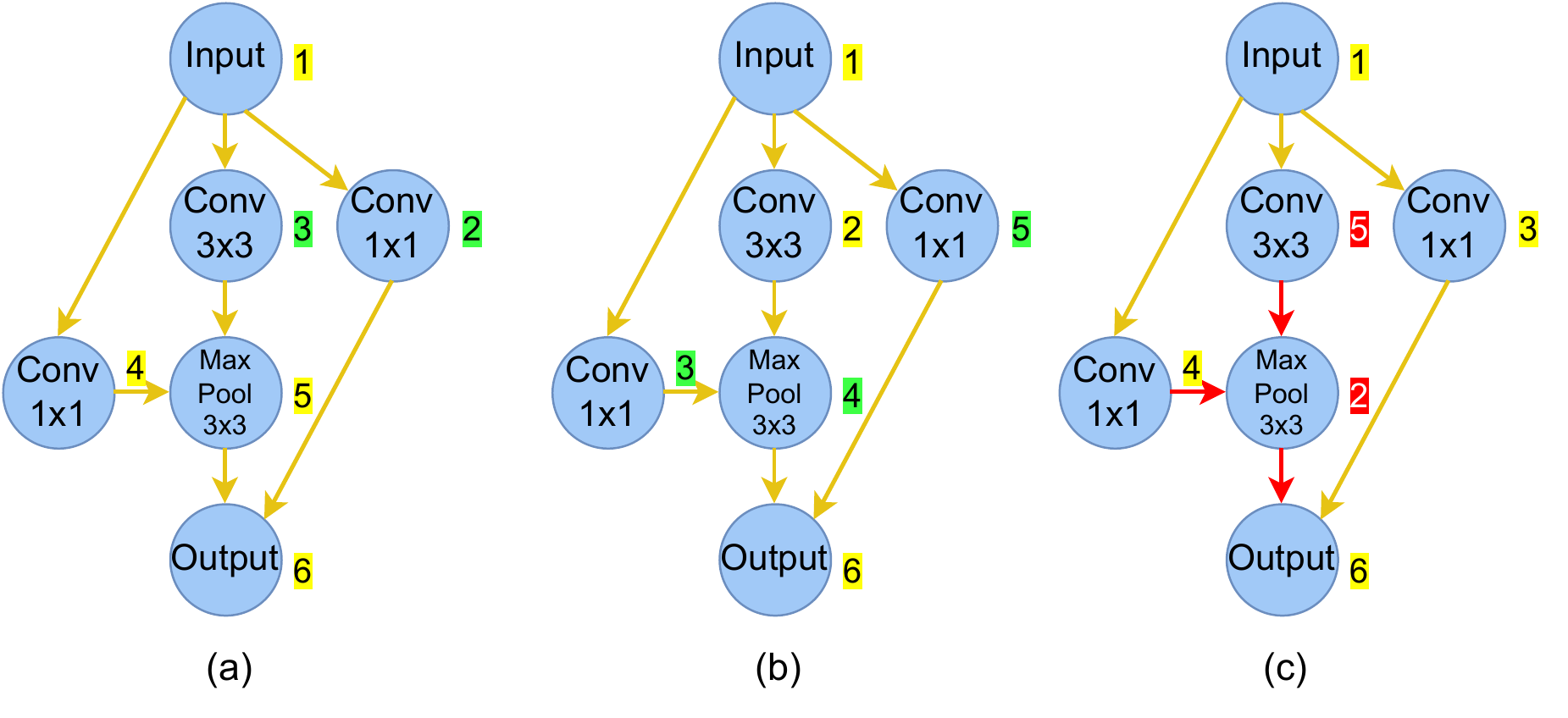}
    \caption{Isomorphic examples of \cref{fig:embedding}(a). (a) and (b) are augmented architectures that maintain the consistency of information flow, while (c) is not.}
    \label{fig:ac_aug}
\end{figure}

The amount of training architecture-attribute data pairs has a significant impact on the effectiveness of our model. However, these datasets are usually small because these attributes of architectures are usually expensive to acquire. For the above reasons, we propose a simple but effective data augmentation method.

For a given architecture with $N$ operations, we can label all operations from $1$ to $N$ (the numbers to the right of operations in \cref{fig:ac_aug}), which indicates the order in which the operations are encoded. Such two different graphs are isomorphic\cite{xie2022architecture} if they have the same node attributes and node connections, only differ in the order of nodes. For instance, architectures in \cref{fig:ac_aug} and \cref{fig:embedding}(a) are isomorphic. 
Recent works \cite{xie2022architecture, liu2021homogeneous} use the idea of isomorphism to augment, and use all isomorphic architectures as augmented data. However, we argue that the prior knowledge that the information flows from shallow to deep nodes facilitates the learning of more accurate representations. Therefore, the architecture with edges flowing from deep nodes to shallow nodes as shown in \cref{fig:ac_aug}(c) is discarded in our augmentation to maintain the information flow consistency between the augmented structure and the original structure.


\subsection{Loss Function}\label{sec:loss}
The total loss function is as follows:
\begin{equation}
    \mathrm{Loss} = \mathrm{MSE\_loss} + \lambda_1 \mathrm{SR\_loss} + \lambda_2 \mathrm{AC\_loss},
\end{equation}
where AC\_loss is the proposed architecture consistency loss, SR\_loss is a loss related to sequence ranking. $\lambda_1$, $\lambda_2$ are the weight coefficients. $\lambda_2$ is equal to zero when there is no augmented data.

\textbf{Architecture Consistency Loss.}
In order to make full use of the augmented data, we propose an architecture consistency loss(AC\_loss) accordingly. It is obvious that the structure obtained by augmentation is equivalent to the original one in terms of operation and topology information. For this reason, we propose such a loss term that their attribute prediction results should be consistent:
\begin{equation}
    \mathrm{AC\_loss}(\widehat{y}, \widehat{y'}) = \sum_{i=1}^M|\widehat{y_i}-\widehat{y'_i}|.
\end{equation}

To train models for the down stream attributes prediction task, we use the Mean Squared Error(MSE) between the predicted results and the true attributes to guide the predictor to generate accurate predictions:
\begin{equation}
    \mathrm{MSE\_loss}(y, \widehat{y}) = \sum_{i=1}^M(\widehat{y}_i - y_i)^2.
\end{equation}

For tasks such as NAS, the prediction accuracy of architectures' relative ranking is more important than that of absolute attribute value. Based on this observation, we introduce a loss related to sequence ranking:
\begin{equation}
    \mathrm{SR\_loss}(y, \widehat{y}) = \sum_{i=1}^M[(\widehat{y}_{I(i)}-\widehat{y}_i)-(y_{I(i)}-y_i)].
\end{equation}
$I(\centerdot)$ is the result of random shuffling of sequences $1$ to $M$.
Note that when architecture augmentation is used, the augmented architectures will also participate in the calculation of MSE\_loss and SR\_loss.
\section{Experiments}
In \cref{sec:nasbench101} and \cref{sec:nasbench201}, we apply our NAR-Former first on NAS-Bench-101\cite{ying2019bench} and NAS-Bench-201\cite{dong2020bench} to demonstrate the superiority of our representation in accuracy prediction tasks. Then in \cref{sec:NAS_Darts} and \cref{sec:NAS_mb}, experiments are conducted to explore the effect of using our predictor on the efficiency and performance of NAS. A small-scale latency prediction experiment is used to verify the generality of the proposed representation model as shown in \cref{sec:NNLQP}. We conduct ablation studies in \cref{sec:ablation} to show the effeciveness of our model design choices.

\subsection{Implementation Details.}
For the tokenizers, we fix $L$ for  $f_{\mathrm{op}}, f_{\mathrm{self}}, f_{\mathrm{sour}}$ all at 32 in \cref{sec:nasbench101}, \cref{sec:nasbench201}, and \cref{sec:NAS_Darts}. 
For settings in \cref{sec:MS-fusion}, we use 6 standard transformer blocks to get $H$ and don't use standard transformer blocks during the fusion stage. We fix the ratio of the hidden dimension to the input dimension in the feed forward networks of transformers to 1:4 and the dimension of each head in multi-head attention to 32. $N_1$, $N_2$ and $N_3$ in \cref{fig:multi-stage} are fixed to 4, 2, and 1, respectively.
$\lambda_1$ is fixed to 0.1 for accurate and latency prediction and 0 for other experiments. $\lambda_2$ is set to 0.5 in \cref{sec:nasbench101} and \cref{sec:nasbench201}. In section \cref{sec:NAS_Darts}, \cref{sec:NAS_mb}, and \cref{sec:NNLQP}, $\lambda 2$ is equal to 0 because augmentation is not used. 
More details about implementation are provide in the supplementary material.
\begin{table}[ht]
    \centering
    \caption{Results on NAS-Bench-101\cite{ying2019bench}(depth=2$\thicksim$7). Kendall's Tau is calculated using predicted accuracies and ground-truth accuracies. Three different proportions of the whole dataset are used as the training set. ``SE'' refer to self-evolution proposed by TNASP\cite{lu2021tnasp} to improve prediction performance.}
    \label{tab:nasbench101}
    \scalebox{0.85}{
    \begin{small}
    \begin{tabular}{llcccc}
    \toprule
    \multirow{5}*{Backbone}    & \multirow{5}{2cm}{Method}              & \multicolumn{4}{c}{Training Samples} \\
           ~                   & ~                                      & 0.1\%  & 0.1\%  & 1\%    & 5\% \\
           ~                   & ~                                      & (424)  & (424)  & (4236) & (21180) \\
    \cline{3-6}
           ~                   & ~                                      & \multicolumn{4}{c}{Test Samples} \\
           ~                   & ~                                      &  100        &  all      & all       &   all      \\
                              
    \hline
    CNN                        & ReNAS\cite{xu2021renas}                & 0.634       & 0.657     & 0.816   & -          \\
    \hline
    
    \multirow{2}*{LSTM}        & NAO\cite{luo2018neural}                & 0.704       & 0.666     & 0.775   & -          \\
                               & NAO+SE                                 & 0.732       & 0.680     & 0.787   & -          \\
    \hline
    \multirow{3}*{GCN}         & NP\cite{wen2020neural}   & 0.710       & 0.679     & 0.769   & -          \\
                               & NP + SE                  & 0.713       & 0.684     & 0.773   & -          \\
                               & CTNAS\cite{chen2021contrastive}        & 0.751       &    -      & -       & -          \\                           
    \hline
    \multirow{3}*{Transformer}  & TNASP\cite{lu2021tnasp}               & 0.752       & 0.705     & 0.820   & -           \\
                                & TNASP + SE                            & 0.754       & 0.722     & 0.820   & -           \\
                                & NAR-Former                            & \textbf{0.801}          & \textbf{0.765}    & \textbf{0.871}            & \textbf{0.891} \\
    \bottomrule
    \end{tabular}
    \end{small}
    }
\end{table}

\subsection{Experiments on NAS-Bench-101}\label{sec:nasbench101}
\noindent{\textbf{Setup.}} NAS-Bench-101\cite{ying2019bench} is a ``operation on nodes'' (OON) space consisting of 423,624 unique architectures. In this experiment, we use Kendall’s Tau \cite{sen1968estimates} as the evaluation metric, and a higher value indicates that the relative order of the predicted values is more related to that of the ground-truth values. 
Following the settings in \cite{lu2021tnasp}, we use 0.1\%, 1\% and 50\% of the whole data as training set. Another 200 samples are used as validation set. 

\begin{table}[ht]
    \centering
    \caption{Results of on NAS-Bench-201\cite{dong2020bench}(depth=8). Kendall's Tau is calculated using predicted accuracies and ground-truth accuracies. Three different proportions of the whole dataset are used as the training set. ``SE'' refer to self-evolution proposed by TNASP\cite{lu2021tnasp} to improve prediction performance.}
    \label{tab:nasbench201}
    \tabcolsep=1em
    \scalebox{0.85}{
    \begin{small}
    \begin{tabular}{llccc}
    \toprule
    \multirow{3}*{Backbone}    & \multirow{3}{2cm}{Model}                   & \multicolumn{3}{c}{Training Samples} \\
           ~                   & ~                                          & (781) & (1563) & (7812) \\
           ~                   & ~                                          & 5\%   & 10\%   & 50\% \\
    \hline
    \multirow{2}*{LSTM}        & NAO\cite{luo2018neural}                    & 0.522    &   0.526    & -          \\
                               & NAO + SE                                   & 0.529    &   0.528    & -          \\
    \hline
    \multirow{2}*{GCN}         & NP\cite{wen2020neural}       & 0.634    &   0.646    & -          \\
                               & NP + SE                      & 0.652    &   0.649    & -          \\
    \hline
    \multirow{3}*{Transformer}  & TNASP\cite{lu2021tnasp}                   & 0.689    &   0.724    & -           \\
                                & TNASP + SE                                & 0.690    &   0.726    & -           \\
                                & NAR-Former                         & \textbf{0.849} & \textbf{0.901} & \textbf{0.947} \\
    \bottomrule
    \end{tabular}
    \end{small}
    }
\end{table}

\noindent{\textbf{Result Comparisons.}}
The results are shown in \cref{tab:nasbench101}, where the values except for our method are reported by TNASP\cite{lu2021tnasp}. When the amount of available training samples is very small, e.g. with only 424, the more recent graph-based methods\cite{chen2021contrastive}\cite{wen2020neural} can obtain higher score than sequence-based methods\cite{luo2018neural}. When more samples are available for training, CNN-based method\cite{xu2021renas} performs better. However, results of above methods are not as good as those of TNASP\cite{lu2021tnasp} in all settings, which confirms the superiority of transformer models in performance prediction. 
By making better use of transformer's capacity and leveraging proposed training strategies, our NAR-Former achieves much higher metrics compared to TNASP\cite{lu2021tnasp}, with an average improvement of 0.047 in the above three settings.


\subsection{Experiments on NAS-Bench-201}\label{sec:nasbench201}
\noindent{\textbf{Setup.}} NAS-Bench-201\cite{dong2020bench} has 15,625 different cell candidates, all of which are composed of 4 nodes and 6 edges. Unlike NAS-Bench-101\cite{ying2019bench}, architectures in NAS-Bench-201\cite{dong2020bench} have its operations on the edges, which belong to ``operation on edge'' (OOE) search space. This change of emphasis has no impact to our implementation, since only the operation types and connection relations are needed to learn the representation. The evaluation indicators is Kendall's Tau\cite{sen1968estimates}. 
The dataset is still divided according to \cite{lu2021tnasp}, where 5\%, 10\% and 50\% of the whole dataset is used for training, and another 200 samples are used for validation. Kendall's Tau is calculated on the whole dataset.

\noindent{\textbf{Results Comparison.}}
 As shown in \cref{tab:nasbench201}, whose values except for our method are reported by TNASP\cite{lu2021tnasp}, earlier method\cite{luo2018neural} using LSTM for encoding has the ability to roughly predict accuracy. Compared with NAO\cite{luo2018neural}, there is a marked increase in the results of the GCN-based method\cite{wen2020neural}. Taking advantage of transformer
 , TNASP\cite{lu2021tnasp} makes a further improvement. 
As we can see, our NAR-Former significantly outperforms by 0.159, 0.175 compared to the previous highest TNASP+SE\cite{lu2021tnasp} under two different experiment settings, respectively. This provides evidence that NAR-Former can learn pretty effective representation of architectures for
accuracy prediction.
\begin{table}[ht]
    \centering
    \caption{Performance of searched architectures using different NAS algorithms in DARTS\cite{liu2018darts} space on CIFAR-10\cite{krizhevsky2009learning}. $\dag$ denotes using cutout\cite{devries2017improved} as data augmentation.}
    \label{tab:NAS}
    \tabcolsep=0.55em
    \scalebox{0.9}{
    \begin{small}
    \begin{tabular}{lcccc}
    \toprule
    \multirow{2}*{Model} & Params & Top1  & No. of  & Search \\
        ~                & (M)                    & Acc(\%)   & archs & Cost(G$\cdot$D)\\
    \hline
    VGG-19~\cite{zhu2021gradinit}              & 20.0    & 95.10           &        0   &  0     \\
    DenseNet-BC\cite{huang2017densely}         & 25.6    & 96.54           &        0   &  0    \\
    Swin-S\cite{liu2021swin}                   & 50      & 94.17           &        0   &  0     \\ 
    Nest-S~\cite{zhang2022nested}              & 38      & 96.97           &        0   &  0     \\
    \hline
    Ransom search                                   & 3.2     & 96.71           &        -      &  -  \\
    NASNet-A$^\dag$ \cite{zoph2016neural}         & 3.3     & 97.35           &       20000   & 1800  \\
    AmoebaNet-A$^\dag$ \cite{real2019regularized}   & 3.2     & 96.66           &       27000   & 3150 \\ 
    PNAS\cite{liu2018progressive}                   & 3.2     & 96.59           &       1160    &  225 \\ 
    NAONet\cite{luo2018neural}                      & 28.6    & 97.02           &       1000    &  - \\
    GATES$^\dag$ \cite{ning2020generic}             & 4.1     & 97.42           &       800     &  - \\ 
    ENAS$^\dag$  \cite{pham2018efficient}           & 4.6     & 97.11           &        -      &  0.5 \\   
    DARTS$^\dag$ \cite{liu2018darts}                & 3.4     & 97.24           &        -      &  4 \\
    CTNAS$^\dag$ \cite{chen2021contrastive}         & 3.6     & 97.41           &        -      &  0.3 \\    
    \hline
    TNASP$^\dag$\cite{lu2021tnasp}                   & 3.7     & 97.48                &       1000   &  0.3    \\
    NAR-Former$^\dag$                                & 3.8     & \textbf{97.52}       &       \textbf{100} & 0.24\\
    \bottomrule
    \end{tabular}
    \end{small}
    }
\end{table}

\begin{table}[ht]
\centering
\caption{Average cost for one sample of NAS-Bench-101\cite{ying2019bench}.}
\tabcolsep=1em
\scalebox{0.9}{
\begin{small}
\begin{tabular}{lccc}
\toprule
Model                          & Encode(ms) & Infer(ms) & Total(ms)  \\
\hline
TNASP\cite{lu2021tnasp}        & 0.0001   & 1.1030    &  1.1031    \\
NAR-Former                     & 2.4784   & 6.4747    &  8.9531 \\
\hline 
\end{tabular}
\end{small}
}
\label{tab:cost}
\end{table}

\begin{table*}[ht]
    \centering
    \caption{Ablation study. The ``Self\_ID'' represents the way in which the self position information is combined with other information. The number in parentheses indicates the amount of augmented data. 
    All experiments follow the setting of 1\% proportion in \cref{sec:nasbench101}.}
    \label{tab:ablation}
    \renewcommand\arraystretch{0.81}
    \begin{small}
    \begin{tabular}{ccccccccc}
    \toprule
     Row   & Architencture           &  Predictor    &  Self\_ID &  SR\_loss  &  GI Aug\cite{xie2022architecture}  &   Our Aug  & AC\_loss & Kendall's    \\
           & Encoder                    & Type           &          &  &  (+3812) &   (+2421 ) &           &    Tau       \\ 
    \hline
     1     & TNASP\cite{lu2021tnasp} & Transformer in \cite{lu2021tnasp}  & -         &   -       &     -    &   -        &  -        &  0.8200   \\
     2     & NP \cite{wen2020neural} & GCN in \cite{wen2020neural}  & -         &   -       &     -    &   -        &  -        & 0.7694  \\
     \hline
     3     & Tokenizer & Transformer in \cite{lu2021tnasp}                & Add       &   -          &     -    &   -        &  -        &  0.8416   \\
     4     & Tokenizer & Transformer in \cite{lu2021tnasp}                & Concat    &   -          &     -    &   -        &  -        &  0.8477   \\
     5     & Tokenizer & GCN in \cite{wen2020neural} & Concat   &   -          &     -    &   -        &  -        &  0.7953    \\ 
     6     & Tokenizer & Multi-stage fusion     & Concat  &   -          &     -    &   -        &  -        &  0.8481   \\
     \hline
     7     & Tokenizer & GCN in \cite{wen2020neural} & Concat &   -          &     -        &   \checkmark  &  -           &  0.8035    \\ 
     8     & Tokenizer & GCN in \cite{wen2020neural} & Concat &   -          &    -         &   \checkmark  &  \checkmark  &  0.8060    \\ 
     
     9     & Tokenizer & Multi-stage fusion  & Concat    &  \checkmark  &     -        &   -           &  -           &  0.8495   \\
     10    & Tokenizer & Multi-stage fusion  & Concat    &  \checkmark  &  \checkmark  &   -           &  -           &  0.8625   \\
     11    & Tokenizer & Multi-stage fusion  & Concat    &  \checkmark  &  \checkmark  &   -           &  \checkmark  &  0.8643   \\
     12     & Tokenizer & Multi-stage fusion  & Concat   &  \checkmark  &     -        &   \checkmark  &  -           &  0.8579   \\
     13     & Tokenizer & Multi-stage fusion  & Concat   &  \checkmark  &     -        &   \checkmark  &  \checkmark  &  0.8712   \\ 
    
    \bottomrule
    \end{tabular}
    \end{small}
\end{table*}

\begin{table}[ht]
    \centering
    \caption{Latency prediction on NNLQP\cite{liu2022nnlqp}. ``Test Model'' denotes the model type that used as test set.}
    \label{tab:latency}
    \tabcolsep=0.9em
    \renewcommand\arraystretch{0.8}
    \scalebox{0.9}{
    \begin{small}
    \begin{tabular}{llll}
    \toprule
    Test Model                          & Method                            & MAPE$\downarrow$     & ACC(10\%)$\uparrow$           \\
    \hline
    ~                                   & FLOPs                             & 58.36\%              &  0.05\%              \\
    EfficientNet                        & TPU\cite{kaufman2021learned}      & \textbf{16.74\%}     & 17.00\%               \\
    depth=242                           & NNLQP\cite{liu2022nnlqp}          & 21.33\%              & \textbf{24.65\% }    \\
    ~                                   & NAR-Former                        & 28.05\%              & 24.08\%             \\
    \midrule
    
    ~                                   & FLOPs                             & 80.41\%              & 0.00\%               \\
    Nas-Bench-201                         & TPU\cite{kaufman2021learned}      & 58.94\%              & 2.50\%               \\
    depth=112$\thicksim$247             & NNLQP\cite{liu2022nnlqp}          & 8.76\%               & 67.10\%                \\
    ~                                   & NAR-Former                        & \textbf{4.19\%}      & \textbf{95.12\%}       \\
    \bottomrule
    \end{tabular}
    \end{small}
    }
\end{table}

\begin{table}[ht]
\centering
\caption{Verification of predictor’s effectiveness using neural structure search experiments on MobileNet space.}
\tabcolsep=1.3em 
\renewcommand\arraystretch{0.8}
\scalebox{0.9}{
\begin{small}
\begin{tabular}{lccc}
\toprule
\multirow{2}*{Model}     & ImageNet   & \multirow{2}*{MACs} & Search Cost  \\
~                        & Top1(\%)   & ~                   & (GPU hours) \\
\hline
OFA\cite{cai2019once}    & 76.00   & 230M    &  40   \\
NAR-Former               & 76.36   & 378M    &  1.63   \\
NAR-Former               & 76.90   & 571M    &  2.00   \\

\hline 
\end{tabular}
\end{small}
}
\label{tab:mobilenet}
\end{table}

\subsection{Experiments on DARTS} \label{sec:NAS_Darts}
\noindent{\textbf{Setup.}} DARTS\cite{liu2018darts} is an open-domain search space, where each architecture consists of normal and reduced cells with 7 nodes and 14 edges. We use three metrics, which are the number of parameters of the searched architecture, the test accuracy on CIFAR-10\cite{krizhevsky2009learning} and the number of structures queried during the search. We use an evolutionary algorithm following NPENAS\cite{wei2022npenas}. Firstly, we randomly sample 10 architectures and train them for 50 epochs on CIFAR-10 as the initial architecture pool. At each round of evolution, we use our predictor that is trained with the structure-accuracy pairs in the structure pool to select the top-10 from the candidate structures generated by mutation, and then the selected structures are added to the structure pool after being trained for 50 epochs. The above steps are repeated until a total of 100 structures are queried. Among these 100 architectures, we choose the architecture with the highest accuracy on validation set to retrain and test.

\noindent{\textbf{Results Comparison.}}
Results of this part are shown in \cref{tab:NAS}. For predictor-based methods, The search cost does not include the time to train the supernet or train the queried architecture to obtain the ground truth.  
Since 2020, the methods adopting graph-based predictor have achieved better results than previous methods, such as 97.41\% for CTNAS\cite{chen2021contrastive} and 97.42\% for GATES\cite{ning2020generic}. The transformer-based predictor TNASP\cite{lu2021tnasp} further improved the accuracy to 97.48\%. 
Taking advantage of our predictor, we continue to improve the accuracy slightly. Compared to \cite{lu2021tnasp}, although our method has a disadvantage in the speed of a single run as shown in \cref{tab:cost}, only one-tenth of the queried architectures help us reduce our search time by 20\%.

\subsection{Experiments on MobileNet Space} \label{sec:NAS_mb}
With the exception of the performance predictor being replaced with our NAR-Former, we strictly follow OFA's\cite{cai2019once} search and evaluation settings. As shown in \cref{tab:mobilenet}, it takes much less time to search for two architectures of different sizes than OFA, and the Top1 accuracy on ImageNet\cite{deng2009imagenet} is still improved by 0.36\% and 0.9\% respectively in the mobile setting (\textless600M MACs).

\subsection{Experiments on NNLQP}\label{sec:NNLQP}
\noindent{\textbf{Setup.}}
NNLQP\cite{liu2022nnlqp} is a latency dataset built for evaluating the performance of latency prediction models. We use the ``unseen structure'' part of this dataset to conduct latency prediction experiments, which contains variants of 9 state-of-the-art DNN models(\eg ResNet\cite{he2016deep}) and NAS-Bench-201\cite{dong2020bench} model, with 2000 different architectures in each model type. The metrics used in this part are Mean Absolute Percentage Error (MAPE) and Error Bound Accuracy (Acc($\delta$)), which 
are the same as used in \cite{liu2022nnlqp}.

\noindent{\textbf{Result Comparison.}}
As can be seen from \cref{tab:latency}, our method can be applied for both the EfficientNet\cite{tan2019efficientnet} and its variants with depths of 242 and the NAS-Bench-201 networks\cite{dong2020bench} with depths ranging from 112 to 247. Combined with the experiments above, these experiment reflect that our model can be used to model structures of very different lengths and to predict different attributes. 


\subsection{Ablation Study}\label{sec:ablation}

We conduct a series of accuracy prediction experiments to verify our contributions following the setting of 1\% proportion in \cref{sec:nasbench101}. The results are shown in \cref{tab:ablation}. The experiments of Row(3)-Row(6) aim to find the optimal structure. By encoding the architecture into a pure sequence with mapping function \cref{eq:pe} and seamlessly using transformer, we achieve a significant improvement (0.0216) compared to the similar transformer-based baseline\cite{lu2021tnasp}(Row(3) vs Row(1)). We argue that, compared with directly adding self position encoding in traditional way, the concatenation of self position information and source position information introduced in \cref{sec:NAE} shows a superior representation of topology information(See Row(3) and Row(4)). A comparison between Row(5) and Row(2) shows that the tokenizer we designed is also applicable to the GCN-based approach.

We explore other strategies to further improve performance, shown as Row(7)-Row(13). For more accurate prediction of relative ranking, we propose SR\_loss, which precisely measures the difference between the predicted ordering and the ground-truth ordering of two inputs. The results in Row(9) and Row(6) show that it works. By the comparison with Row(9), the results in Row(10) and Row(12) demonstrate that both augmentation methods are useful. The data set generated by our AC\_Aug is a subset of that generated by GIAug. The additional data generated by GI\_Aug can be seen as a secondary augmentation, which brings more benefit(0.0130) than AC\_Aug(0.0084) when only using data augmentation. The extra augmentation samples can not keep the consistency of information flow, making it less advantageous in obtaining performance improvement when using AC\_loss(Row(11) and Row(13)). The results in  Row(7) and Row(8), compared with Row(5), also verify the effectiveness of the proposed AC\_Aug and AC\_loss on the GCN-based model, respectively.

\section{Conclusion}
We propose an effective neural architecture representation learning framework that are consisted of linearly scaling network encoders, a transformers based representation learning model, and an effective model training method with data augmentations and assisted loss functions. Experiments show that our framework are capable of improving the accuracy of downstream prediction tasks while overcoming scale limitations on input architectures. Although not the scope of this work, we believe this framework can also be extended for other down stream tasks, such as predicting the quantization loss or searching for the best mixed precision model inference strategies. 

\section*{Acknowledgement}
This work was supported in part by the National Key Research and Development Program of China under Grant 2018AAA0103202; in part by the National Natural Science Foundation of China under Grants U22A2096 and 62036007; in part by the Technology Innovation Leading Program of Shaanxi under Grant 2022QFY01-15; in part by Open Research Projects of Zhejiang Lab under Grant 2021KG0AB01.

\clearpage

{\small
\bibliographystyle{ieee_fullname}
\bibliography{egbib}
}

\end{document}